\documentclass[runningheads]{llncs}
\usepackage[T1]{fontenc}
\usepackage{graphicx}

\usepackage{subfig}
\usepackage{algorithm}
\usepackage{multirow}
\usepackage[noend]{algpseudocode}
\usepackage{listings}
\usepackage{hhline}
\usepackage{adjustbox}
\usepackage{tablefootnote}
\usepackage[
  font = small,
  labelfont = bf,
  tableposition = top
]{caption}

\usepackage{color}
\begin{document}

\title{Automatic evaluation of herding behavior\\ in towed fishing gear using end-to-end training\\ of CNN and attention-based networks}
\titlerunning{Automatic evaluation of herding behavior using end-to-end training}
%
\author{Orri Steinn Guðfinnsson \and Týr Vilhjálmsson \and\\ Martin Eineborg \and Torfi Thorhallsson }
\authorrunning{O.\ S.\ Guðfinnsson, T. Vilhjálmsson, M. Eineborg, and T. Thorhallsson}

%
\institute{Department of Engineering, Reykjavik University, Iceland \\
\email{\{orri17,tyr17,martine,torfith\}@ru.is}}

\maketitle              
\begin{abstract}
    This paper considers the automatic classification of herding behavior in the cluttered low-visibility environment that typically surrounds towed fishing gear. The paper compares three convolutional and attention-based deep action recognition network architectures trained end-to-end on a small set of video sequences captured by a remotely controlled camera and classified by an expert in fishing technology.
    The sequences depict a scene in front of a fishing trawl where the conventional herding mechanism has been replaced by directed laser light. The goal is to detect the presence of a fish in the sequence and classify whether or not the fish reacts to the lasers. A two-stream CNN model, a CNN-transformer hybrid, and a pure transformer model were trained end-to-end to achieve  63\%, 54\%, and 60\% 10-fold classification accuracy on the three-class task when compared to the human expert.   
    Inspection of the activation maps learned by the three networks raises questions about the attributes of the sequences the models may be learning, specifically whether changes in viewpoint introduced by human camera operators that affect the position of laser lines in the video frames may interfere with the classification. This underlines the importance of careful experimental design when capturing scientific data for automatic end-to-end evaluation and the usefulness of inspecting the trained models.
    
\end{abstract}

\section{Introduction}

Deep networks have been successfully employed to detect human actions in short video sequences without explicit representation of the actors or objects \cite{simonyan_two-stream_2014}. The models are trained end-to-end by providing only a single action label for each sequence. No explicit representations, such as objects, locations, or tracks, are extracted or provided. Instead, the model learns an implicit representation from the data and the associated action labels during training. This promises certain optimality of the representation concerning the imagery and the action classes, side-steps failure modes of intermediate object detection or tracking algorithms, and eliminating the manual labor of annotating every frame in a sequence. On the downside, the resulting model is essentially a black box and raises the question of what information in the images the model uses as a base for its predictions.

Results reported in the literature are typically obtained by training on a large set of video sequences sourced from the internet (e.g., \ UCF-101 \cite{UCF101} consisting of 13 320 video clips and 101 action classes). Although varied, the sequences generally depict reasonably framed, well-lit scenes with the quality expected from modern color video cameras.

In contrast, underwater imagery used in fisheries research is often captured in low-visibility environments using monochrome cameras and may be further restricted by using ambient illumination for unobtrusive observation.

The number of available samples can also be limited.  
However, in many deep learning applications, good results have been reported on small datasets using transfer learning, i.e., \ by fine-tuning a model pre-trained on a large dataset.

This paper investigates how state-of-the-art deep action recognition networks perform when trained end-to-end on a relatively small set of underwater video sequences.
The dataset consists of underwater videos of fish obtained in experimental fishing trials, along with action labels encoding a behavior observed by an expert in fishing technology. 

The fact that the models do not explicitly predict the location of the fish in each frame means that the reasoning behind the predicted class is not readily verified. In an attempt to elucidate the reasoning, the activation maps \cite{gradcam} learned by these networks are inspected for evidence that the model is attentive to the area where the fish is observed.

After introducing the dataset and the action recognition algorithms, the performance of the trained models is presented. This is followed by an inspection of the trained models using activation maps. Finally, the difference between observed and expected activation sources is discussed and further explored to identify possible sources of bias in the dataset that could adversely affect the accuracy of the model predictions.

\section{Background}

\subsection{Related work}
Although there exists substantial literature proposing novel deep learning methods for tasks using underwater imagery (e.g., \ \cite{JALAL2020101088,novelmethod,MALOY2019105087}), there is not much that offers insight into the difficulties associated with working with data of this kind. Most underwater imagery used in deep learning features scenes of shallow water ecosystems or footage captured close to the ocean surface. As such, these scenes do not suffer from problems accompanied by bad lighting conditions or noisy backgrounds. While reading through the available literature, we could not find any works using video data captured using a moving camera featuring deep seabed scenes similar to ours. \cite{MALOY2019105087} uses a dual-stream network for action classification and deals with high clutter video data. However, the videos are captured in an aquaculture environment, with the fish behaving more predictably, lacking distinct poses, as the fish mostly swim in a circle and only pass the camera moving from right to left. In \cite{RAHMAN2014574} the authors use a novel action descriptor and fish action data to classify fish movement. They also compare their results to state-of-the-art methods. Their dataset is comparable in size to ours with 95 videos but only contains shallow water scenes.

\subsection{Annotated video of fish behaviour}\label{sec:AnnotatedVideo} 

\begin{figure}[b]
\centering
\begin{tabular}{llll}
\centering
 \includegraphics[width=.2\linewidth]{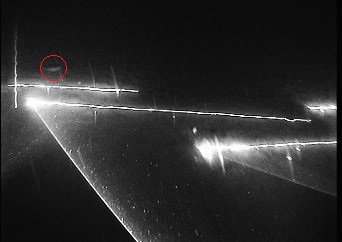} & \includegraphics[width=.2\linewidth]{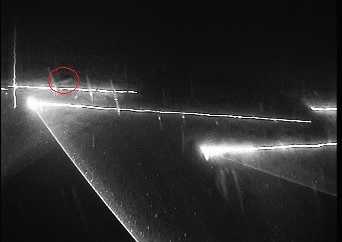} & \includegraphics[width=.2\linewidth]{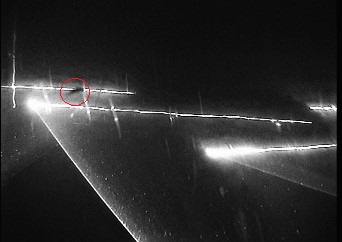} & \includegraphics[width=.2\linewidth]{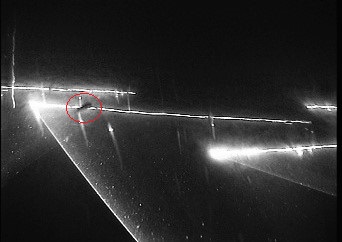}\\
 \includegraphics[width=.2\linewidth]{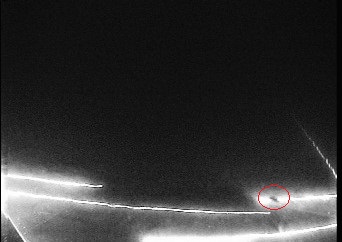} & \includegraphics[width=.2\linewidth]{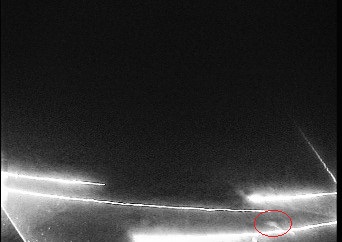} & \includegraphics[width=.2\linewidth]{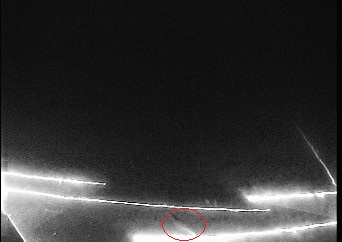} &
 \includegraphics[width=.2\linewidth]{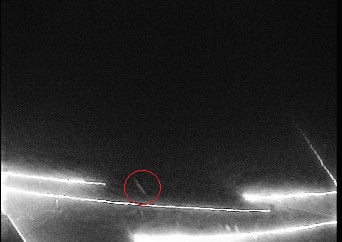}\\

\end{tabular}
\captionof{figure}{Example frames from the underwater videos. The top row shows an example from the No Reaction (NR) class, while the bottom row shows an example from the Reaction (R) class. A red circle marks the location of the fish.\newline}
\label{fig:ExampleVideoFrames}
\end{figure}

The dataset used in this paper consists of underwater videos  \cite{thorhallsson_tr2018_trawling} recorded from a camera mounted on an experimental trawl, viewing the area immediately in front of the trawl. The trawl is unique in that a virtual structure projected by lasers has replaced the forward section of the trawl, as can be seen in Fig. \ref{fig:ExampleVideoFrames}.
The recordings were made to evaluate the effectiveness of herding fish using directed light.
The work reported here examines if the evaluation can be automated, allowing an evaluation to run continuously over extended trials without human effort.

The fishing trials \cite{thorhallsson_tr2018_trawling} were conducted in April 2018 off the Snæfellsnes peninsula, W-Iceland, at a water depth between 80 and 120 meters, where daylight is strongly attenuated. The observations span two hauls conducted on separate days totaling two hours.
An enhanced CCD low-light monochrome camera (Kongsberg oe15-100c) was mounted on the trawl via a remotely controllable pan-tilt head to allow fishing gear specialists to observe the fish behavior in real-time. No additional camera lighting was used in order not to affect the behavior.

The analog video stream was digitized to 8-bit depth and stored frame-by-frame in JPEG-compressed image format at a high-quality setting.
Post-operation, the videos were played back at half speed by an expert who registered the behavior of each observed fish as showing either no response, weak response, or strong response together with a time-stamp.

\subsection{Two-stream convolutional networks}
The Two-stream Convolutional Network \cite{simonyan_two-stream_2014} is a neural network architecture proposed for action recognition in videos. The network is composed of two parallel Convolutional Neural Networks (CNNs): a spatial stream and a temporal stream that each produces as output a prediction of the action class.  

The input to the spatial stream is a single video frame, while the input to the temporal stream is a stack of adjacent optical flow images.
The output of the two streams is then fused by adding the softmaxed class score. Note that the two streams are trained separately; thus, the output of one stream does not affect the other.

\subsection{Transformers}

Transformer networks \cite{vaswani2017attention} were originally proposed as an alternative to recurrent neural networks used in natural language processing (NLP). Transformers learn to attend to prior words in a sequence, providing a significant improvement on NLP problems while drastically reducing training time. 
More recently, transformers have been adapted to various computer vision tasks \cite{pmlr-v139-bertasius21a,Arnab_2021_ICCV}. In the Vision Transformer \cite{dosovitskiy2021an} a pure transformer is used for image classification by dividing each image down into a sequence of patches and using attention mechanisms to capture relationships between different areas of an image.
Transformer architectures can be divided into two categories: hybrid transformers, where the transformer architecture is accompanied by a recurrent or a convolutional network, and a pure transformer, where the architecture is based only on attention mechanisms.
Transformers have been similarly adapted to video classification tasks by applying the attention mechanism to sequences of frames. Two recent examples are the "TimeSformer" \cite{pmlr-v139-bertasius21a} and "ViViT" \cite{Arnab_2021_ICCV}, that both have demonstrated competitive results on multiple video classification benchmarks.

\subsection{Visualizing Activation Maps}

Gradient-weighted Class Activation Mapping or Grad-CAM \cite{gradcam} is a method that uses the gradients of a target class to produce a map that indicates the regions of an input image that are most important for predicting the target class. These activation maps can be used to explain failure modes (causes of apparently unreasonable predictions), identify dataset biases, and provide an intuitive approach for tuning a model.

\section{Methods}
\subsection{Dataset}

\subsubsection{Scene complexity}
Before training the models on the dataset, we observed some aspects of the dataset that might introduce difficulties for the models. We list a few of them below.

\begin{itemize}
    \item The videos have a moving background that often contains objects that are easily mistaken for fish.
    \item The scene is dark, and the fish have natural camouflage that matches the seabed.
    \item The videos contain lasers which complicate the background further.
    \item The videos contain a time-stamp.
    \item Interpreting and classifying fish actions for data labeling is not straightforward, which results in samples of various lengths. (Every fish response is different.)
    \item The fishes are most often small compared to the scene.
\end{itemize}

Given the facts mentioned above and the fact that deep learning architectures have a tendency to use flaws found in a dataset to their advantage when trying to find features that separate samples of different classes, it is essential to understand how a model interprets the data it is being fed so as to avoid misinterpreting the results. Good performance is meaningless if the correct conclusion is derived from flawed arguments.

\subsubsection{Dataset generation}

For the work presented here, short video clips of around 1-second duration (approx.\ $30$ frames) were extracted from the video recordings (Sec. \ref{sec:AnnotatedVideo}) around each registered observation and sorted by the time of capture.

For experimentation, the weak reaction samples were omitted from the dataset, and samples were added where no fish was visible. Each video clip in the resulting dataset was thus labeled as belonging to one of three classes: Reaction (R), No Reaction (NR), and No Fish (NF).
Example frames from two classes can be seen in Fig.\ \ref{fig:ExampleVideoFrames}. 
The distribution of the sequence length (Fig.\ \ref{fig:DatasetFrameCount}) is similar across the three classes.

The dataset was split into training, validation, and testing sets for training and validation.
For a more accurate statistical evaluation, a total of 10 random splits were generated.
The number of clips of each class in each set can be seen in Table \ref{tab:DatasetVideos}.

As the viewing direction of the camera can be controlled by a human operator, the camera field of view is not identical in every sample in the dataset but has some variation, as can be seen in Fig. \ref{fig:ExampleVideoFrames}.\newline

\noindent
\begin{minipage}{\textwidth}
  \begin{minipage}[b]{0.49\textwidth}
    \centering
    \includegraphics[width=5cm]{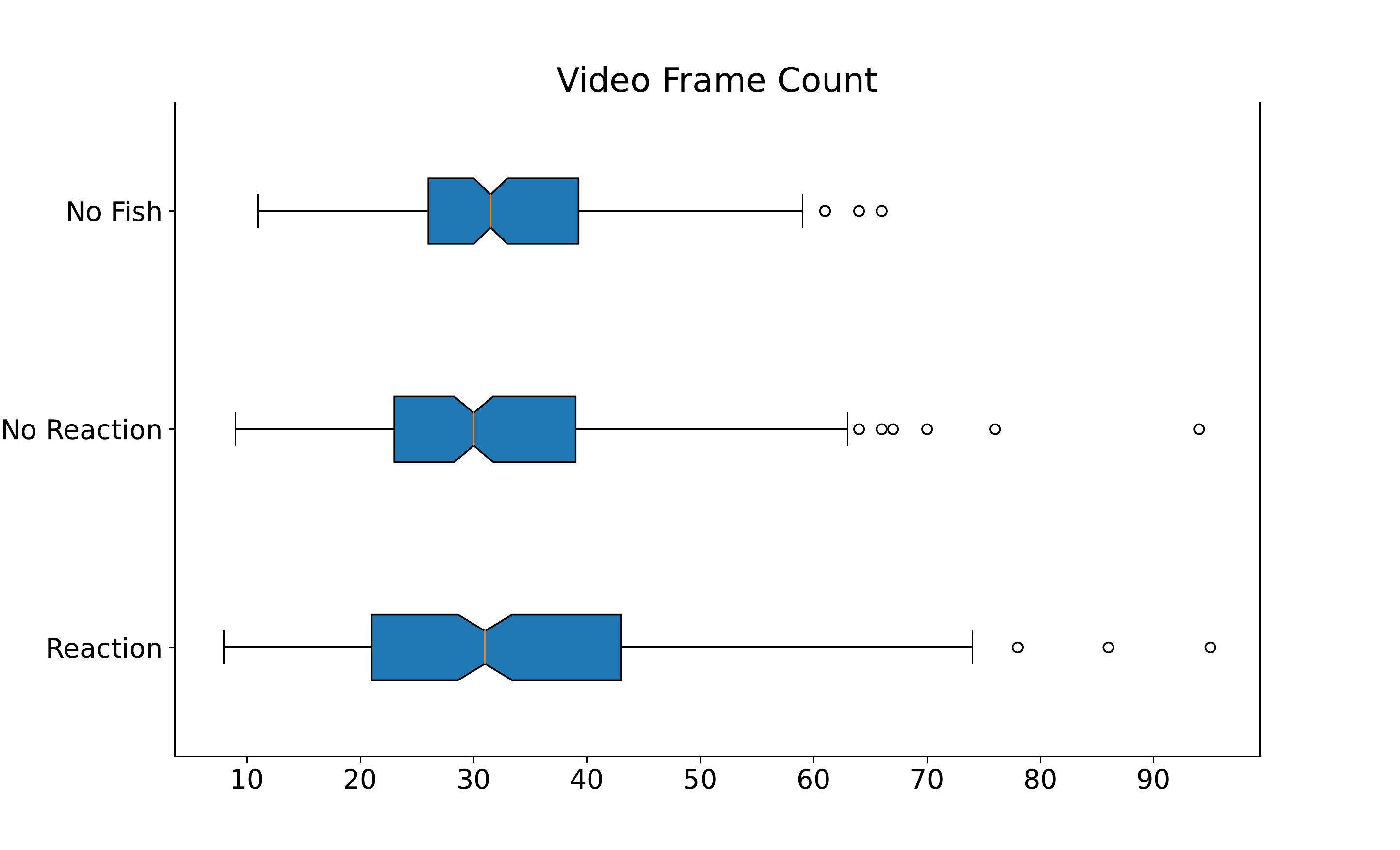}
    \captionof{figure}{The frame count distribution is shown as a box-plot for each of the three classes in the dataset.}
    \label{fig:DatasetFrameCount}
  \end{minipage}
  \hfill
  \begin{minipage}[b]{0.49\textwidth}
    \centering
    \captionof{table}{The class distributions for the data splits. All ten splits contained the same number of samples from each class, i.e., all ten training sets contained 144 "No fish" samples, all validation sets contained 39 "No Reaction" samples, and so on.}
\begin{tabular}[b]{|c|c|c|c|c|} 
\hline
Class       & Training & Validation & Testing & Total  \\ 
\hline
NF    & 144          & 36             & 20          & 200    \\ 
\hline
NR & 154          & 39             & 21          & 214    \\ 
\hline
R    & 151          & 38             & 21          & 210    \\ 
\hhline{|=====|}
Total       & 449          & 113            & 62          & 624    \\
\hline
\end{tabular}
\label{tab:DatasetVideos}
  \end{minipage}
\end{minipage}

\subsection{Data pre-processing}

Before training, the input images were cropped to exclude a time-stamp imprinted in the top left corner of each frame.

The optical flow images used as an input to the temporal stream in the two-stream network were created using the Farnebäck method \cite{Farneback}. The method's input parameters were adjusted slightly by experimentation to obtain a more accurate motion estimate from the videos.

\subsection{Model architectures}
\noindent
The two-stream network used here differs from the original \cite{simonyan_two-stream_2014}
in that, we sample eight uniformly distributed frames from a video, send them individually into the spatial stream, and average the classification output for each frame. For the input to the temporal stream, instead of stacking the x- and y-components of the optical flow (as discussed in \cite{simonyan_two-stream_2014}), we compute the total optical flow and stack that with the original frames in the video.
The architecture used for both streams in the two-stream network is a resnet18  CNN \cite{resnet}. The hybrid model was built using a resnet18 as a feature extractor, a VisionTransformer encoder with five layers and five heads, and a classifier consisting of an AdaptiveMaxPool1d layer, a dropout layer, and finally, a fully connected layer. The timeSformer model used the base configuration of visionTransformer with divided space-time attention. The patch size used for timeSformer was 16x16.

\subsection{Training procedure}

The hyperparameters used when training the models are shown in table \ref{tab:Hyperparameters}. Hyperparameter optimization was conducted using grid search. A range of values were tested for each hyperparameter. 

\begin{table}[h]
\centering
\caption{Hyperparameters in training}
\begin{tabular}{|c|c|c|c|c|c|} 
\hline
Model       & Learning rate & Epochs & Batch size & Image size & Frames/Video  \\ 
\hline
Spatial CNN     & $1e-4$          & 200    & 4         & 300x300          & 8   \\ 
\hline
Temporal CNN & $1e-4$          & 200       & 4      & 300x300          & 2x7    \\ 
\hline
Hybrid    & $1e-6$          & 100        & 4     & 300x300          & 12   \\ 
\hline
TimeSformer       & $1e-6$          & 100    & 3        & 224x224          & 8   \\
\hline
\end{tabular}
\label{tab:Hyperparameters}
\end{table}

Using more frames per video and a larger image size has been shown to improve performance for video classification tasks \cite{dosovitskiy2021an}. However, due to the large number of parameters these models have and the resulting memory consumption, we choose our parameters by balancing performance and training time.

Transformer architectures have also been shown to benefit significantly from pretraining due to the same reasons \cite{pmlr-v139-bertasius21a}. Both the two-stream and hybrid models used a resnet18 pre-trained on ImageNet \cite{imagenet}. The hybrid encoder was trained from scratch as pretrained weights were not readily available. The timeSformer used a VisionTransformer, pretrained on ImageNet. The final model for each split was chosen using early stopping, based on cross-entropy loss values computed on the corresponding validation set during training.

The spatial CNN and the TimeSformer network used a random horizontal flip for data augmentation. No data augmentation was used for the temporal CNN and the hybrid transformer model. Note that for the spatial and temporal CNNs, a learning rate scheduler was used where the learning rate decreases by a factor of 10 if the model does not improve for ten epochs.

\section{Experiments}
Three network models were trained to predict fish behavior using the training dataset. 
Section 4.1 presents a statistical evaluation of the performance of the trained models using common performance measures.
Sections 4.2 and 4.3 provide observations on the image attributes used by the models based on the Grad-CAM activation maps together with the predicted probability (PP) of individual classes. 
The PP is computed by applying the softmax function to the model output, which puts the output values for all the classes on the interval (0,1) and the sum of the values equal to 1. Then the values can be interpreted as probabilities. The general softmax function is given by the formula

\begin{equation}
    \label{eq:confidence}
    \sigma(x_i) = \frac{e^{x_i}}{\sum_0^{N_c-1}e^{x_i}}
\end{equation}
where $x_i$ is the model output for class, i, and $N_c$ is the number of classes.

\subsection{Comparing model performance}

\noindent
All three models were tuned on a single split, split 0, before being trained on splits 1 through 10 and evaluated on the test sets. Two metrics were used to evaluate the performance of each model, the overall accuracy and the F1 score for the NF class. The F1 score, given by equation \ref{eq:f1score}, is used to quantify each model's ability to recognize the difference between a video containing a fish and a video that does not.

\begin{equation}
\label{eq:f1score}
    F_1 = 2\times \frac{precision\times recall}{precision + recall} = 2\times \frac{tp}{tp+\frac{1}{2}(fp+fn)}
\end{equation}

\begin{table}[h]
\centering
\caption{Comparison of the accuracy achieved by each model on splits 1-10. The mean accuracy for each model across all splits is highlighted in the last column, along with the respective standard deviation.}
\begin{adjustbox}{width=1\textwidth}
\begin{tabular}{l l l l l l l l l l l | l} 
\hline
\textbf{Data split} & \textbf{1} & \textbf{2} & \textbf{3} & \textbf{4} & \textbf{5} & \textbf{6} & \textbf{7} & \textbf{8} & \textbf{9} & \textbf{10} &   \\
\hline
\textbf{Two-Stream} &67.74&64.52&61.29&59.68&59.68&62.90&70.97&66.13&54.84&66.13& \textbf{63.39 $\pm$ 4.45}\\
Spatial  & 56.45 & 66.13 & 58.06 & 58.06 & 58.06 & 67.74 & 58.06 & 67.74 & 56.45 & 54.84 & \textbf{60.16 $\pm$ 4.73} \\
Temporal &67.74&62.90& 56.45 &56.45&54.84&59.68&69.35&69.35&53.23&64.52 &\textbf{61.45 $\pm$ 5.83}\\
\textbf{Hybrid} & 56.45 & 45.16 & 59.68 & 53.23 & 48.39 & 56.45 & 56.45 & 50.00 & 48.39 & 61.29 & \textbf{53.55 $\pm$ 5.09}\\
\textbf{TimeSformer} & 64.52&	62.90	&62.90	&58.06	&54.84	&61.29	&59.68	&54.84	&62.90	&61.29	&\textbf{60.32 $\pm$ 2.77}\\\hline

\end{tabular}
\end{adjustbox}
\label{Accuracy}
\end{table}

Table \ref{Accuracy} shows the 10-fold accuracy achieved by each model. As shown in the table, the two-stream network achieves the highest accuracy. Of the transformer models, the pure transformer model outperforms the hybrid model. Interestingly, the best performing splits for the hybrid model seem to be the worst for the timeSformer. This trend is much less apparent in the F1 scores, as seen in table \ref{f1score}, mainly because the performance gap is significantly increased.

\begin{table}[h]
\centering
\caption{Comparison of the F1 score achieved by each model on splits 1-10 for the NF class. The mean F1 score for each model across all splits is highlighted in the last column, along with the respective standard deviation.}
\begin{adjustbox}{width=1\textwidth}
\begin{tabular}{l l l l l l l l l l l | l} 
\hline
\textbf{Data split} & \textbf{1} & \textbf{2} & \textbf{3} & \textbf{4} & \textbf{5} & \textbf{6} & \textbf{7} & \textbf{8} & \textbf{9} & \textbf{10} &  \\
\hline
\textbf{Two-Stream} &78.26&78.26&70.59&70.83&73.08&71.43&80.85&71.70&68.00&73.17&\textbf{73.62 $\pm$ 3.91}\\
Spatial  & 72.22 & 68.57 & 82.93 & 70.83 & 68.00 & 82.05 & 70.27 & 71.43 & 63.16 & 63.64 & \textbf{71.31 $\pm$ 6.29}\\
Temporal &78.26&73.91&64.15&66.67&69.23&69.77&76.00&76.00&67.92&76.19 & \textbf{71.81 $\pm$ 4.60}\\
\textbf{Hybrid} & 70.59 & 46.15 & 66.67 & 61.54 & 54.55 & 53.33 & 63.16 & 54.90 & 54.55 & 64.00 & \textbf{58.94 $\pm$ 7.03}\\
\textbf{TimeSformer} & 78.26	&73.68&	69.57	&68.18	&66.67&	66.67&	68.18&	75.00	&74.42	&75.00 & \textbf{71.56 $\pm$ 3.71}\\\hline
\end{tabular}
\end{adjustbox}
\label{f1score}
\end{table}

\subsection{Class Activation Maps}
For both the two-stream network and the hybrid network, the class activation maps targeted the final convolutional block in the resnet18. For the timeSformer model, we used the norm1 layer in the last encoder block.

\subsubsection{Laser line positions affected by the camera view}

We looked at the Grad-CAMs from every model for all the test splits, and they all showed a similar trend. Consistently, the most activity was on the lasers and the neighboring pixels, as clearly shown in Figs. \ref{fig:looklasers} and \ref{fig:fishnlaser}. There are some exceptions to this, but generally, this was the case.

Figure \ref{fig:looklasers} shows a Grad-CAM example from test set 2. In this case, the TimeSformer correctly predicted the video as belonging to class R with a relatively high confidence of $75.15\%$. The fish is clearly seen in the frame, yet according to the Grad-CAM, the model does not deem it an important factor in determining the class of the video. This is not a unique example; in fact, most if not all of the Grad-CAMs observed showed this pattern in some capacity. There are cases, such as the one shown in Fig. \ref{fig:fishnlaser}, where the model also considers the fish, but we assume that the reason for that is that the fish is apparent in common laser area.\\

\noindent
\begin{minipage}{\textwidth}
  \begin{minipage}[b]{0.49\textwidth}
    \centering

    \includegraphics[width=6cm]{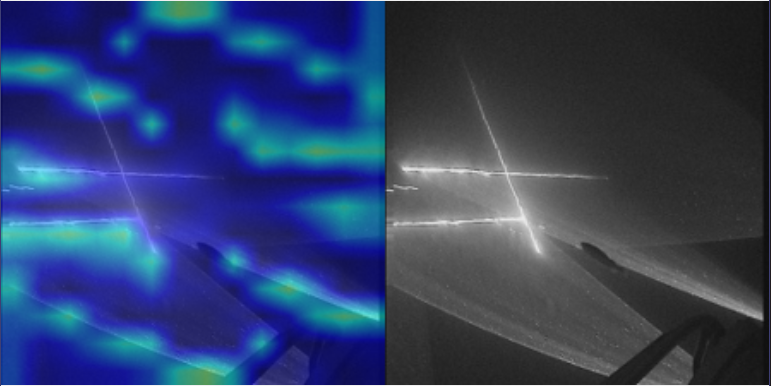}
    \captionof{figure}{The Grad-CAM (left) shows activation mostly occurs around the laser, with no activation around the fish, even though the fish is clearly visible in the original frame (right) as it covers a portion of the laser, close to the center of the image. The model still classifies the video correctly and with high confidence.\\}
    \label{fig:looklasers}

  \end{minipage}
  \hfill
  \begin{minipage}[b]{0.49\textwidth}
    \centering

    \includegraphics[width=6cm]{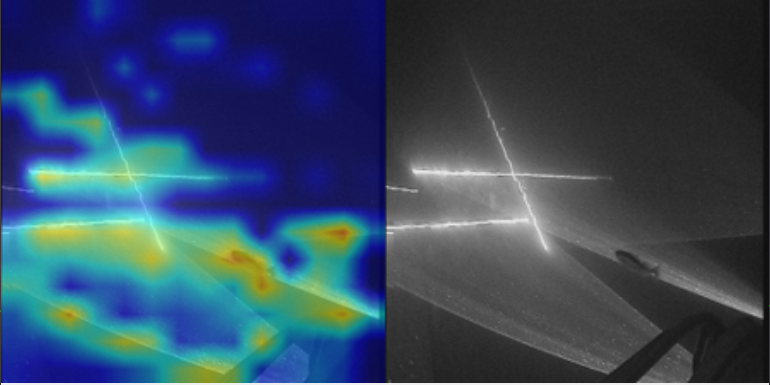}

    \captionof{figure}{The Grad-CAM (left) shows activation around the lasers and the fish. The red color around the fish indicates high activation. However, the activation is likely due to fish covering an area where the model expects the laser to shine brightly.\\\\}
    \label{fig:fishnlaser}

  \end{minipage}
\end{minipage}

\subsubsection{Inclusion of a time-stamp in the frame}

\begin{figure}[h]
    \centering
    \includegraphics[width=.4\textwidth, trim={1.8cm 2.1cm 1.8cm 2.1cm},clip]{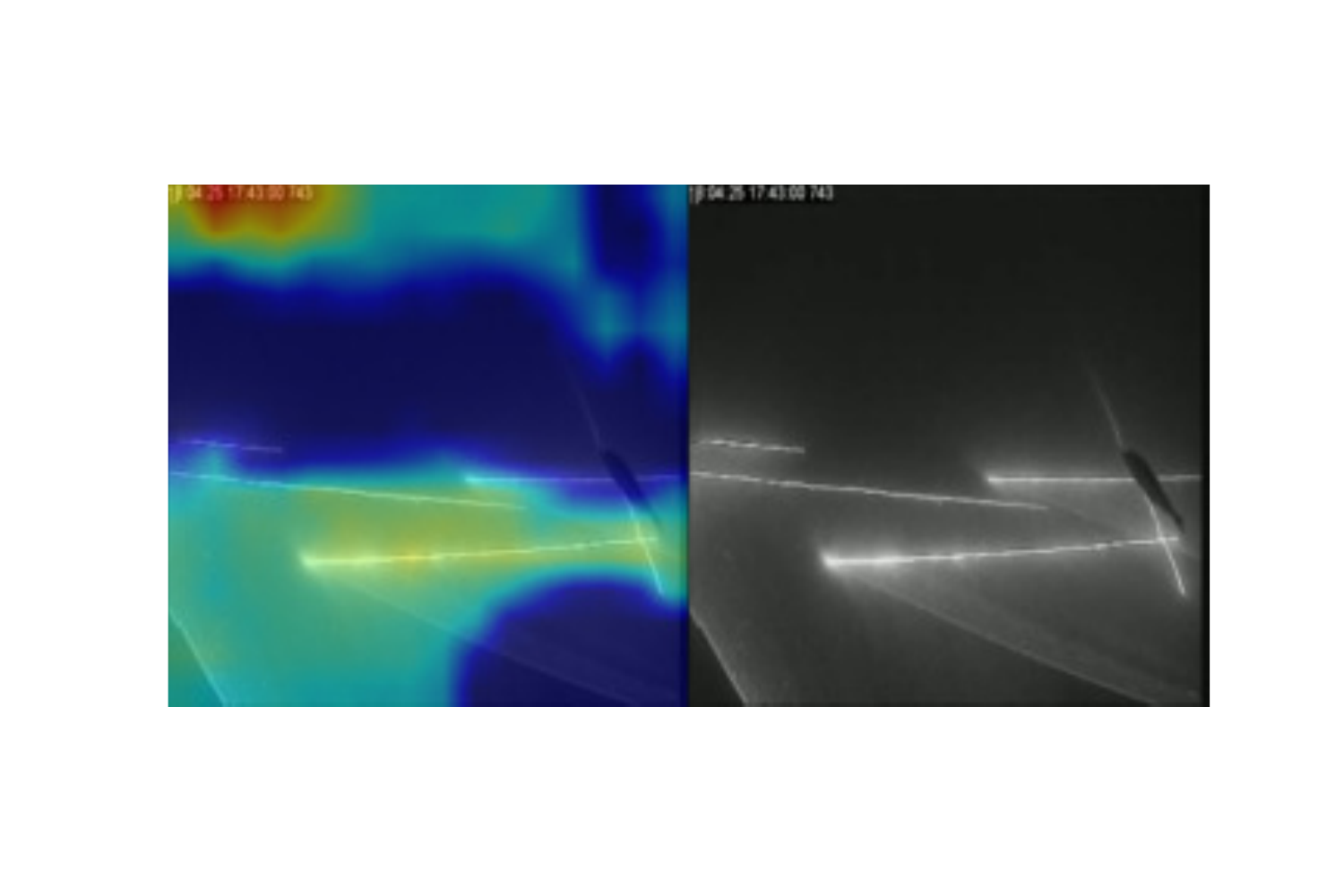}
    \caption{An example of the model's activation around the time-stamp in the top-left corner of the video frame.}
    \label{fig:TimeStamp}
\end{figure}

The original video frames include a time-stamp in the top left corner that was cropped away in pre-processing.
The activation maps of models trained on the uncropped image data (Fig.\ \ref{fig:TimeStamp}) show strong activation around the time-stamp. However, on close inspection, the activation appears to be most strongly focused on the day and the minute within the hour.

This indicates that the model may be basing its prediction on the time of day observed in the training samples.

\subsubsection{Max sequence length - Padding frames}

\begin{figure}[h]
    \centering
    \includegraphics[width=0.2\textwidth]{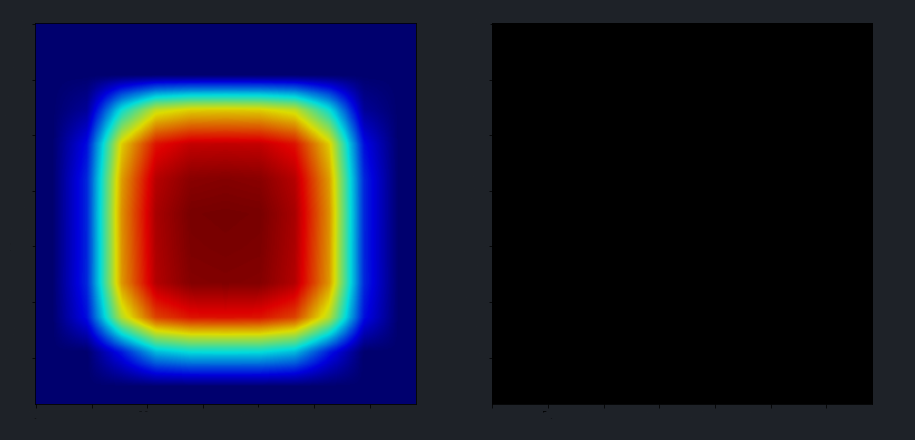}
    \caption{Class activation of a padding frame. In the ideal situation, the Grad-CAM (left) and the original frame (right) would be the same as the Grad-CAM would not show any activation. However, as seen in the Grad-CAM, the padding frame has very high activation, meaning the model deems the empty frame critical to the classification.}
    \label{fig:blackframe}
\end{figure}

\noindent
As previously mentioned, increasing the length of a video sequence has been shown to increase the accuracy of video classification models. The same is valid for increasing the number of data samples for training. The consequence of attempting to achieve both is that the samples can be of different lengths. Coupled with the fact that different actions take different amounts of time to complete, it is very probable that some samples will be longer than others. Since deep learning models cannot take input of variable size, this becomes a problem. The simplest solution for this problem is to limit all video sequences to the length of the shortest video in the dataset.\\
Nevertheless, this can have severe adverse effects if the difference between the longest and shortest video is significant. Another method for combating this problem is to pad shorter videos with zeros up to a certain length, which can also have adverse effects. Figure \ref{fig:blackframe} shows an example of a padding frame used in an experiment with a sequence length of 40, along with the corresponding Grad-CAM. In this experiment, several different sequence lengths were tested to see if performance could be enhanced using data only from class R and NR. Although the performance seemed to improve with a larger sequence length, the observed improvement was due to the model using the number of padding frames to classify class R videos. Generally, when using padding frames, the model should regard the empty frames as irrelevant information and, in turn, ignore them. However, as Fig. \ref{fig:blackframe} indicates, the model is extracting features from the empty frames. The PP for class R videos containing padding frames was also found to be 100\% in all cases. Furthermore, the PP for all videos containing padding frames was also found to be 100\% for class R, i.e., the model always classified videos containing padding frames as class R, with 100\% certainty.

As detailed in Fig. \ref{fig:DatasetFrameCount}, the dataset contains videos as short as eight frames and videos that are over 75 frames long. A closer examination of the average video length for each class shows that the R class contains shorter videos on average compared to the NR class. This is the most likely explanation for the increase in reaction class accuracy as the sequence length increases. 

\subsection{Potential Classification bias}

\begin{figure}[h]
    \centering
    \includegraphics[width=0.9\textwidth]{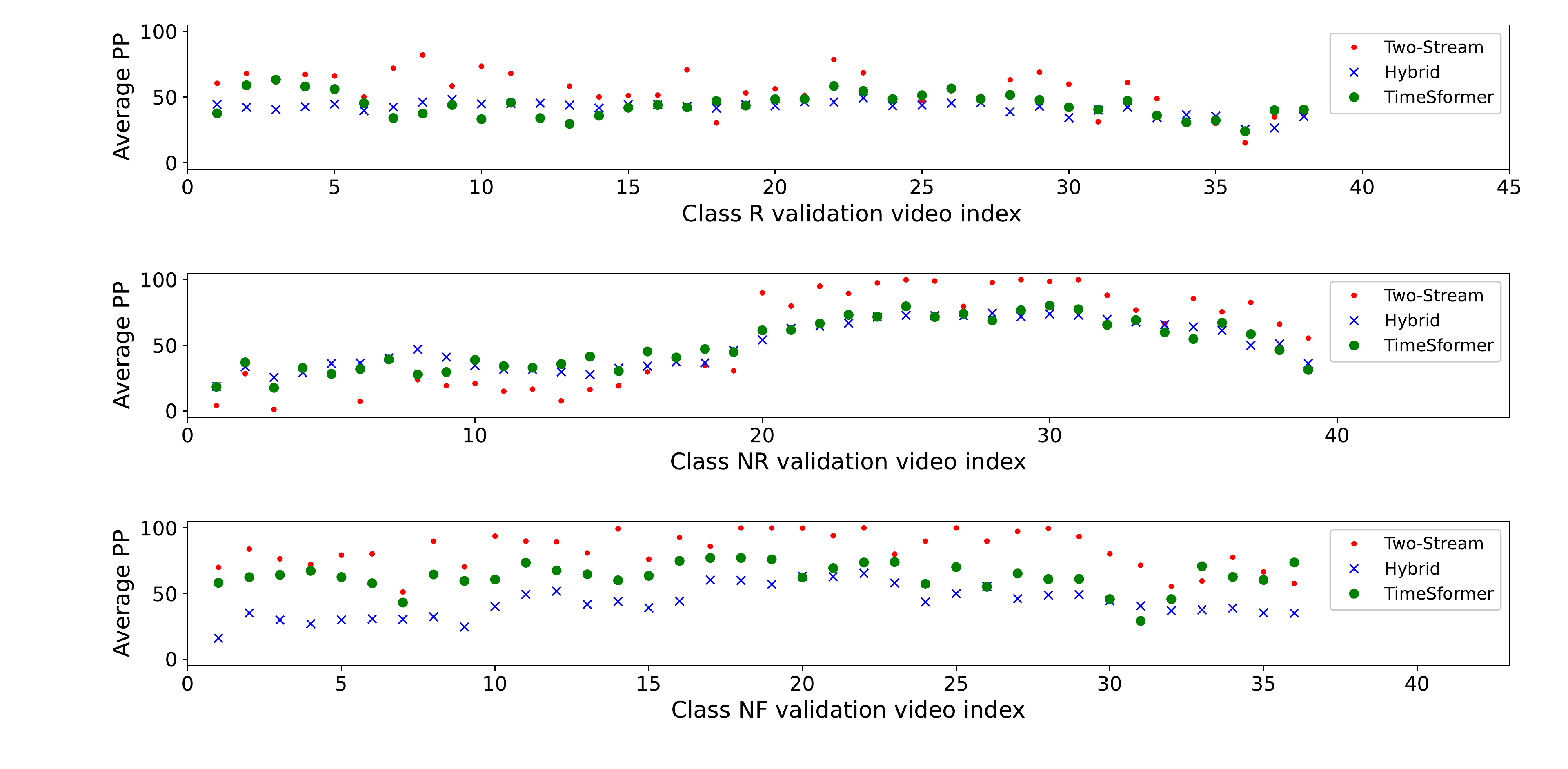}
    \caption{The average predicted probability (PP) for every video in the validation sets based on its class.}
    \label{fig:AverageTestPredProb}
\end{figure}

Figure \ref{fig:AverageTestPredProb} shows the average PP per video across all 10 splits on the validation dataset. The order of the videos in the validation dataset corresponds to the order in which the video clips were created from the raw video footage. The average PP curves from each model follow a similar trend, indicating an underlying bias in the dataset.

\begin{figure}
\centering
\begin{tabular}{llll}
\centering
 \includegraphics[width=.17\linewidth]{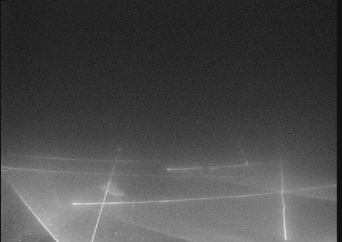} & \includegraphics[width=.17\linewidth]{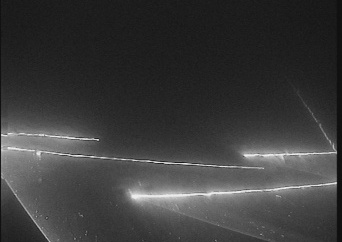} & \includegraphics[width=.17\linewidth]{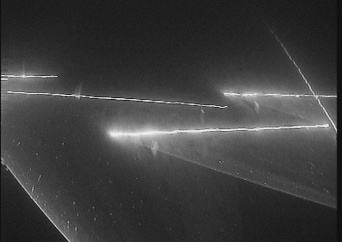} & \includegraphics[width=.17\linewidth]{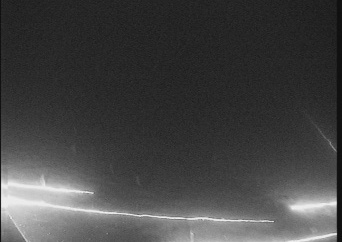}\\
 \includegraphics[width=.17\linewidth]{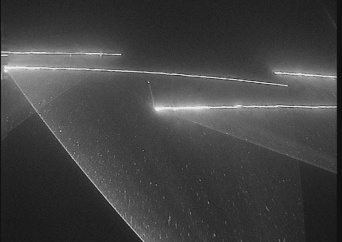} & \includegraphics[width=.17\linewidth]{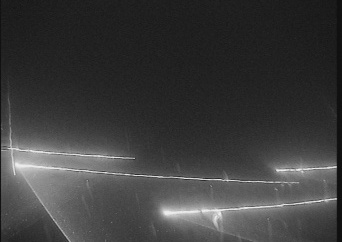} & \includegraphics[width=.17\linewidth]{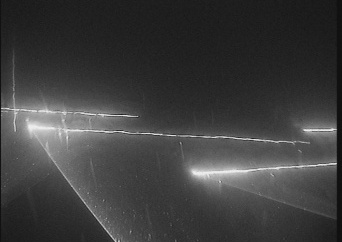} &
 \includegraphics[width=.17\linewidth]{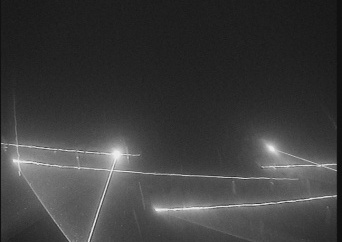}\\
 \includegraphics[width=.17\linewidth]{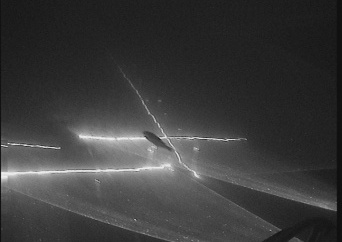}& \includegraphics[width=.17\linewidth]{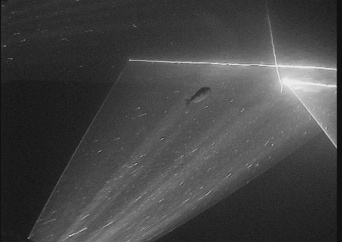} & \includegraphics[width=.17\linewidth]{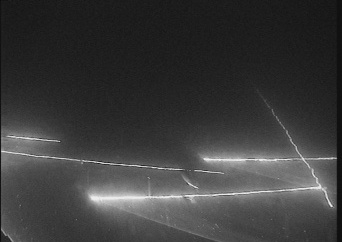} &
 \includegraphics[width=.17\linewidth]{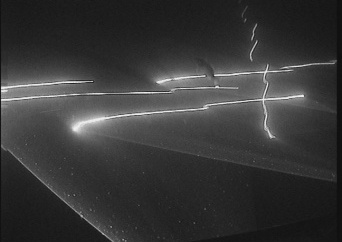}\\
  \includegraphics[width=.17\linewidth]{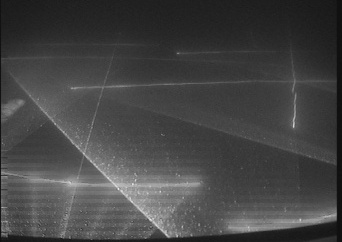} & \includegraphics[width=.17\linewidth]{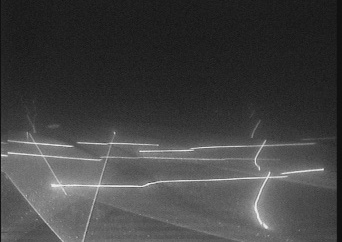} & \includegraphics[width=.17\linewidth]{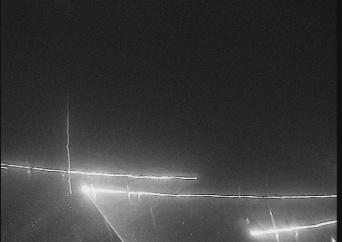} &
 \includegraphics[width=.17\linewidth]{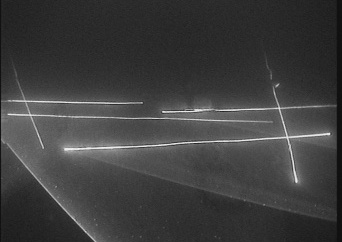}\\
\end{tabular}
\captionof{figure}{Example frames from the 16 different camera views show how the position of the lasers differs between the views. Within each view, the position of the prominent laser lines is subtly modulated by the terrain and the motion of the trawl. }
\label{fig:CameraViews}
%
\end{figure}

As mentioned in Section \ref{sec:AnnotatedVideo}, the camera's viewing direction could be remotely controlled by the camera operator. It is indeed observed that the viewing direction is not the same across all samples, causing the position of the lasers to differ between samples, which could be a source of bias. 
To further examine the possibility of a human-introduced bias, the videos in the dataset were manually grouped into 16 camera views according to viewing direction (Fig.\ \ref{fig:CameraViews}).

\begin{figure}[h]
    \centering
    \includegraphics[width=0.4\textwidth]{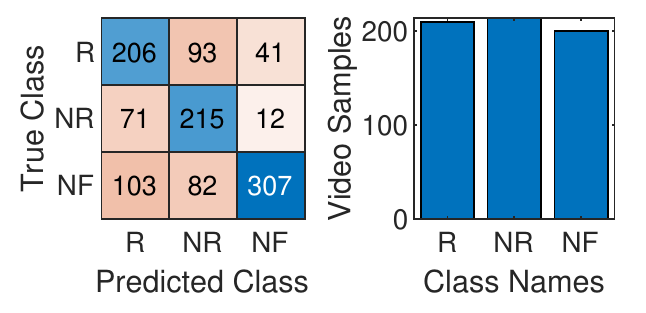}
    \caption{The combined confusion matrix from all validation splits, along with the class distribution in the dataset.}
    \label{fig:TotalConfusionMatrix}
\end{figure}

\begin{figure}[H]
    \centering
    \includegraphics[width=\textwidth]{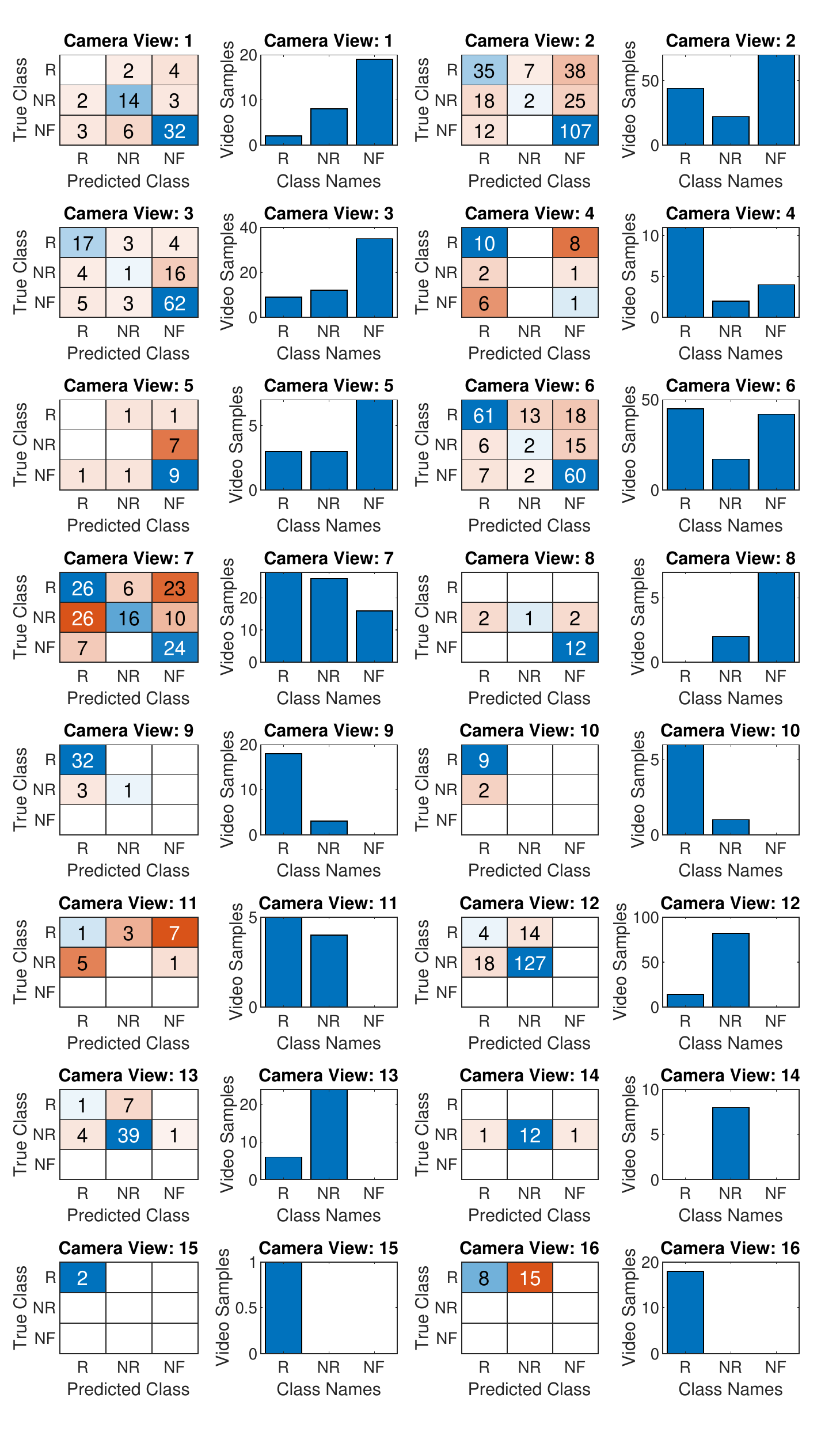}
    \caption{Confusion matrices for each of the 16 views and the distribution of samples in each view.  With few exceptions, the samples are far from being balanced across the three classes.}
    \label{fig:ConfusionMByScenesTwoStream}
\end{figure}

Using the results from the best performing network (two-stream), 
a separate confusion matrix was created for the samples within each camera view. The confusion matrices  (Fig.\ \ref{fig:ConfusionMByScenesTwoStream}) contain the combined predictions across all ten validation sets. The figure also gives the prevalence of classes in each camera view in the form of a bar plot, from which it is apparent that the dataset is far from being balanced within each view. The bar plots accompanying each confusion matrix contain the class distribution for each camera view over the whole dataset. Figure \ref{fig:TotalConfusionMatrix} shows the combined confusion matrix across all ten validation sets, along with the class distribution in the dataset. In contrast to Fig. \ref{fig:ConfusionMByScenesTwoStream}, Fig. \ref{fig:TotalConfusionMatrix} shows that there is little imbalance in the classes in the dataset, when camera views are not taken into account.

As Fig. \ref{fig:ConfusionMByScenesTwoStream} shows, the two-stream network often predicts the most frequently occurring class in each camera view, which could explain its relatively high accuracy on the dataset, despite the Grad-CAMs indicating that the models rarely recognize the fish. Note that the confusion matrices shown in Figs. \ref{fig:TotalConfusionMatrix} and \ref{fig:ConfusionMByScenesTwoStream} only show results from the two-stream network. However, the confusion matrices for the transformer networks were highly similar to the ones shown in the figures.

The models would consistently predict the most prevailing class for most camera views shown in Fig. \ref{fig:ConfusionMByScenesTwoStream}. However, there are cases where the confusion matrices do not follow the class distribution, e.g., camera views 4, 7, 11, and 16, which might indicate subtle differences in the scenes contained within the camera view that is only recognized by the model. Additionally, there are cases (e.g., views 1, 3, and 6) where the model accurately predicts the actual class of the video despite the actual class not being the prevailing class in the distribution. This could also indicate that the model partly learns to look at the fish in the scene.

\section{Conclusions}

The paper compares the performance of three action recognition network architectures on a relatively small dataset. The dataset consists of short underwater video sequences of fish behavior in front of an unconventional trawl. The dataset is challenging due to the videos having a moving background, the fish blending in with the cluttered environment, the lasers apparent in the scene may confuse the networks, and the fish is most often small compared to the scene.
Each model was evaluated on ten random splits, and performance metrics such as the F1 score for the NF class and the total accuracy were recorded. The two-stream network achieved 63.39\% 10-fold accuracy and 73.62\% F1 score, the highest of the tested models.

Grad-CAMs indicate that the models are predominantly attentive to the location of the laser lines in the image. Only rarely a fish also appears in the region of activation, as shown in Fig. \ref{fig:fishnlaser}. This mainly occurs when the fish is in the vicinity of the lasers. Given that the models consistently achieve an accuracy above guess rate, this observation points to an underlying bias between the location of the lasers in the frame and the presence and reaction of the fish. Therefore the classification of adjacent videos was examined since adjacent videos are more likely to have the same camera view. Figure \ref{fig:AverageTestPredProb} clearly shows a trend that the models classify adjacent videos the same. 
By observing the NR subplot, on average, the first half of the NR videos are incorrectly classified by the models, whereas the second half is correctly classified. Similar trends can be found in the other classes, but the trend is most apparent in the NR class. 

To further examine if there is an underlying bias in the dataset, the videos in the dataset were grouped into 16 different camera views, as shown in Fig. \ref{fig:CameraViews}. The confusion matrices for each camera view (Fig. \ref{fig:ConfusionMByScenesTwoStream}) showed that the algorithms mostly predicted the most frequent class for the corresponding camera view. This might explain the relatively high accuracy of the algorithms on the dataset even though the Grad-CAM indicated that the models showed little to no attention to the fish in the frames.

When training end-to-end networks, care must be taken not to introduce a bias into the data being processed. As seen in this work, the models learned the dataset to some extent but not as intended due to bias introduced by human tampering.

\subsection*{Acknowledgements}
This work is partly supported by the Rannís Technology Development Fund under grant number 2010831-0612.

%
%
%
%

\bibliographystyle{IEEEtran}
\bibliography{References}

\begin{thebibliography}{10}
\providecommand{\url}[1]{#1}
\csname url@samestyle\endcsname
\providecommand{\newblock}{\relax}
\providecommand{\bibinfo}[2]{#2}
\providecommand{\BIBentrySTDinterwordspacing}{\spaceskip=0pt\relax}
\providecommand{\BIBentryALTinterwordstretchfactor}{4}
\providecommand{\BIBentryALTinterwordspacing}{\spaceskip=\fontdimen2\font plus
\BIBentryALTinterwordstretchfactor\fontdimen3\font minus
  \fontdimen4\font\relax}
\providecommand{\BIBforeignlanguage}[2]{{%
\expandafter\ifx\csname l@#1\endcsname\relax
\typeout{** WARNING: IEEEtran.bst: No hyphenation pattern has been}%
\typeout{** loaded for the language `#1'. Using the pattern for}%
\typeout{** the default language instead.}%
\else
\language=\csname l@#1\endcsname
\fi
#2}}
\providecommand{\BIBdecl}{\relax}
\BIBdecl

\bibitem{simonyan_two-stream_2014}
K.~Simonyan and A.~Zisserman, ``Two-stream convolutional networks for action
  recognition in videos,'' in \emph{Advances in Neural Information Processing
  Systems}, vol.~27, 2014.

\bibitem{UCF101}
\BIBentryALTinterwordspacing
K.~Soomro, A.~R. Zamir, and M.~Shah, ``{UCF101}: A dataset of 101 human actions
  classes from videos in the wild,'' 2012. [Online]. Available:
  \url{https://arxiv.org/abs/1212.0402}
\BIBentrySTDinterwordspacing

\bibitem{gradcam}
R.~R. Selvaraju, M.~Cogswell, A.~Das, R.~Vedantam, D.~Parikh, and D.~Batra,
  ``Grad-{CAM}: Visual explanations from deep networks via gradient-based
  localization,'' \emph{International Journal of Computer Vision}, vol. 128,
  no.~2, pp. 336--359, oct 2019.

\bibitem{JALAL2020101088}
A.~Jalal, A.~Salman, A.~Mian, M.~Shortis, and F.~Shafait, ``Fish detection and
  species classification in underwater environments using deep learning with
  temporal information,'' \emph{Ecological Informatics}, vol.~57, p. 101088,
  2020.

\bibitem{novelmethod}
S.~Siddiqui, A.~Salman, I.~Malik, F.~Shafait, A.~Mian, M.~Shortis, and
  E.~Harvey, ``Automatic fish species classification in underwater videos:
  Exploiting pretrained deep neural network models to compensate for limited
  labelled data,'' \emph{ICES Journal of Marine Science}, vol.~75, 05 2017.

\bibitem{MALOY2019105087}
H.~Måløy, A.~Aamodt, and E.~Misimi, ``A spatio-temporal recurrent network for
  salmon feeding action recognition from underwater videos in aquaculture,''
  \emph{Computers and Electronics in Agriculture}, vol. 167, p. 105087, 2019.

\bibitem{RAHMAN2014574}
S.~A. Rahman, I.~Song, M.~Leung, I.~Lee, and K.~Lee, ``Fast action recognition
  using negative space features,'' \emph{Expert Systems with Applications},
  vol.~41, no.~2, pp. 574--587, 2014.

\bibitem{thorhallsson_tr2018_trawling}
T.~Thorhallsson, E.~Hreinsson, H.~Karlsson, G.~Gudmundsson, H.~Jónsdóttir,
  and G.~Haney, ``Trawling with light,'' RANNIS Technology Development Fund,
  Project 153487-0613, Technical Final Report, [In Icelandic], Dec 2018.

\bibitem{vaswani2017attention}
A.~Vaswani, N.~Shazeer, N.~Parmar, J.~Uszkoreit, L.~Jones, A.~N. Gomez,
  {\L}.~Kaiser, and I.~Polosukhin, ``Attention is all you need,''
  \emph{Advances in neural information processing systems}, vol.~30, 2017.

\bibitem{pmlr-v139-bertasius21a}
G.~Bertasius, H.~Wang, and L.~Torresani, ``Is space-time attention all you need
  for video understanding?'' in \emph{Proceedings of the 38th International
  Conference on Machine Learning}, ser. Proceedings of Machine Learning
  Research, M.~Meila and T.~Zhang, Eds., vol. 139.\hskip 1em plus 0.5em minus
  0.4em\relax PMLR, 18--24 Jul 2021, pp. 813--824.

\bibitem{Arnab_2021_ICCV}
A.~Arnab, M.~Dehghani, G.~Heigold, C.~Sun, M.~Lu\v{c}i\'c, and C.~Schmid,
  ``Vivit: A video vision transformer,'' in \emph{Proceedings of the IEEE/CVF
  International Conference on Computer Vision (ICCV)}, October 2021, pp.
  6836--6846.

\bibitem{dosovitskiy2021an}
A.~Dosovitskiy, L.~Beyer, A.~Kolesnikov, D.~Weissenborn, X.~Zhai,
  T.~Unterthiner, M.~Dehghani, M.~Minderer, G.~Heigold, S.~Gelly, J.~Uszkoreit,
  and N.~Houlsby, ``An image is worth 16x16 words: Transformers for image
  recognition at scale,'' in \emph{International Conference on Learning
  Representations}, 2021.

\bibitem{Farneback}
G.~Farneb{\"a}ck, ``Two-frame motion estimation based on polynomial
  expansion,'' in \emph{Image Analysis}, J.~Bigun and T.~Gustavsson, Eds.\hskip
  1em plus 0.5em minus 0.4em\relax Berlin, Heidelberg: Springer Berlin
  Heidelberg, 2003, pp. 363--370.

\bibitem{resnet}
K.~He, X.~Zhang, S.~Ren, and J.~Sun, ``Deep residual learning for image
  recognition,'' in \emph{2016 IEEE Conference on Computer Vision and Pattern
  Recognition (CVPR)}, 2016, pp. 770--778.

\bibitem{imagenet}
J.~Deng, W.~Dong, R.~Socher, L.-J. Li, K.~Li, and L.~Fei-Fei, ``Imagenet: A
  large-scale hierarchical image database,'' in \emph{2009 IEEE Conference on
  Computer Vision and Pattern Recognition}, 2009, pp. 248--255.

\end{thebibliography}

\end{document}